%% file: acl2020.tex
\documentclass[11pt,a4paper]{article}
\usepackage[hyperref]{acl2020}
\usepackage{times}
\usepackage{latexsym}

\usepackage{booktabs}
\usepackage{graphicx}
\usepackage{multirow}

\usepackage{microtype}

\aclfinalcopy 

\title{\texttt{MWPToolkit}: An Open-source Framework for Deep Learning-based Math Word Problem Solvers}

\author{
Yihuai Lan\textsuperscript{\rm{1}}\thanks{\ \ Equal contribution.},
	Lei Wang\textsuperscript{\rm{2}}\footnotemark[1],
	Qiyuan Zhang\textsuperscript{\rm{2}\footnotemark[1]}, 
	Yunshi Lan\textsuperscript{\rm{3}}\thanks{\ \ Corresponding author.},
	Bing Tian Dai\textsuperscript{\rm{2}},\\
	\textbf{Yan Wang\textsuperscript{\rm{4}}},
	\textbf{Dongxiang Zhang\textsuperscript{\rm{5}}},
	\textbf{Ee-Peng Lim\textsuperscript{\rm{2}}}\\
	
	\textsuperscript{1} Xihua University, 
	\textsuperscript{2}Singapore Management University \\
	\textsuperscript{3}East China Normal University, 
	\textsuperscript{4} Tencent AI Lab,
	\textsuperscript{5} Zhejiang University\\
	
	\small {\tt \{lei.wang.2019, yslan.2015\}@phdcs.smu.edu.sg,\  \{btdai, eplim\}@smu.edu.sg} \\
	\small {\tt \{yifan2250, qiyuanzhang97\}@gmail.com},\
	\small {\tt bradenwang@tencent.com},\
	\small {\tt zhangdongxiang@zju.edu.cn}
}

\date{}

\begin{document}
\maketitle
\begin{abstract}

Developing automatic Math Word Problem (MWP) solvers has been an interest of NLP researchers since the 1960s.
Over the last few years, there are a growing number of datasets and deep learning-based methods proposed for effectively solving MWPs. 
However, most existing methods are benchmarked solely on one or two datasets, varying in different configurations, which leads to a lack of unified, standardized, fair, and comprehensive comparison between methods. 
This paper presents \texttt{MWPToolkit}, the first open-source framework for solving MWPs. 
In \texttt{MWPToolkit}, we decompose the procedure of existing MWP solvers into multiple core components and decouple their models into highly reusable modules. 
We also provide a hyper-parameter search function to boost the performance. 
In total, we implement and compare 17 MWP solvers on 4 widely-used single equation generation benchmarks and 2 multiple equations generation benchmarks. 
These features enable our \texttt{MWPToolkit} to be suitable for researchers to reproduce advanced baseline models and develop new MWP solvers quickly. 
Code and documents are available at \url{https://github.com/LYH-YF/MWPToolkit}. 

\end{abstract}

\section{Introduction}

Developing automatic Math Word Problems (MWPs) solvers has been an interest of NLP researchers since the 1960s~\cite{feigenbaum1963computers, bobrow1964natural}. 
As shown in Table~\ref{tab:example}, when solving a MWP, machines need to make inferences based on the given textual problem description and question. 
It requires machines to translate the natural language text into valid and solvable equations according to context, numbers, and unknown variables in the text and then compute to obtain the numerical values as the answer.

\input{Tables/example}

Over the last few years, there are a growing number of datasets~\cite{Kushman2014learning, Koncel2016mawps, Upadhyay2017AnnotatingDA,  Huang2016dolphin, wang2017dns, qin2020sau-solver, Miao2020ADC, Patel2021AreNM} and deep learning-based methods that have been proposed to solve MWPs, 
including Seq2Seq~\cite{wang2017dns, wang2018mathen, chiang2019saligned, li2019groupatt}, Seq2Tree~\cite{wang2019trnn, liu2019astdec, xie2019gts, qin2020sau-solver}, Graph2Tree~\cite{zhang2020graph2tree, shen2020multiED}, and Pre-trained Language Models~\cite{kim2020ept}. 
However, most existing MWP solving methods are evaluated solely on one or two datasets, varying
in different settings (i.e., different train-test split and $k$-fold cross validation), which leads to a lack of unified, standardized, fair, and comprehensive comparison between methods. 
Moreover, it is time-consuming and complicated to re-implement prior methods as baselines, which leads to the difficulty making a consistent conclusion in term of performance comparison to other methods.
Thus it severely hinders the development of research in MWPs community. 

To encourage the development of this field, we present \texttt{MWPToolkit}, the first open-source framework for deep learning-based MWP solvers. 
To unify MWP methods into \texttt{MWPToolkit}, we design the framework of MWP solvers as an architecture with multiple core components: config, data, model and evaluation. 
We further decouple the components into highly reusable modules and deploy them into \texttt{MWPToolkit}.
Thus, it is easily extensible and convenient to develop new models by combing existing modules and replacing individual modules with proposed ones. 
Besides, we also develop a hyper-parameter search function for all methods developed in \texttt{MWPToolkit}, which helps mitigate the negative impact caused by sub-optimal hyper-parameters.

\texttt{MWPToolkit} includes comprehensive benchmark datasets and models. 
So far, we have incorporated 6 widely-used MWP datasets and 17 models.
The datasets contain 4 datasets that are single equation generation and 2 datasets that are multiple equation generation.
The models include Seq2seq, Seq2tree, Graph2tree, and commonly-used non-pretrained (AttSeq2Seq~\cite{bahdanau2014neural}, LSTMVAE~\cite{zhang2016lstmvae}, and Transformer~\cite{vaswani2017attention}) and pretrained models (GPT-2~\cite{radford2019gpt2}, BERTGen~\cite{devlin2018bert}, and RoBERTaGen~\cite{liu2019roberta}). 
Currently, our framework supports built-in evaluation protocols including equation accuracy and answer accuracy for two types of generation. 
In our \texttt{MWPToolkit}, users can run, compare, and test models on MWP tasks under the same setting with simple configuration files and command lines.

To ensure that our re-implementations in \texttt{MWPToolkit} are correct and the experiments by our framework are reliable, we set the same hyper-parameters as the ones in original papers and ensure the re-implemented result should be approximate to the reported result.
In this paper, we provide a set of results of 17 models on 6 datasets with the same $k$-fold cross-validation setting  after the built-in hyper-parameter search. 
We hope the community can benefit from the results of this comprehensive comparison, better understand existing MWP methods, and easily develop new and powerful models by utilizing our \texttt{MWPToolkit}.

\section{\texttt{MWPToolkit} Framework}

\begin{figure}[t] 
\centering
\includegraphics[width=\linewidth]{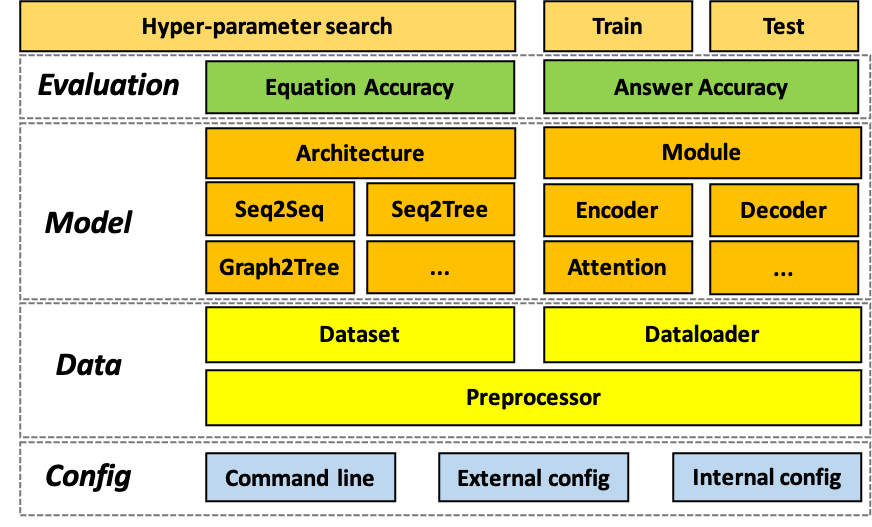} 
\vspace{-15px}
\caption{The overall framework of \texttt{MWPToolkit}.}
\label{fig:framework} 
\vspace{-10px}
\end{figure}

The overall framework of our \texttt{MWPToolkit} is presented in Figure~\ref{fig:framework}, including the config component, data component, model component, evaluation component, and execution from bottom to top.
The config component is used to set up the experimental configuration, which supports the following components. 
The data component preprocess different datasets into a unified form used in the subsequent components. 
The model component is responsible for model construction. 
After determining specific evaluation metrics in the evaluation component, the execution part is used to train with a given group of hyper-parameters or hyper-parameter searching and evaluate models with a specific setting, i.e., train-test split or $k$-fold cross-validation. 
Note that in our \texttt{MWPToolkit}, users can use the given random seed to reproduce results completely. 
In the following part, we present the details of config, data, model, evaluation components and execution part. 

\subsection{Config Component}

Config component serves as the core human-system interaction component in which the developer can specify experimental configurations. 
The configuration part of our framework consists of \textit{Command lines}, \textit{External config} and \textit{Internal config}.
The default configuration is defined in internal configuration file.
Users can flexibly and simply use the command lines to modify the major settings of the experiment.
They can also have more customized settings with external configuration files, which is  beneficial for the duplication of research results.

\subsection{Data Component}

\input{Tables/datasets}

Any raw dataset in the data module follows the pre-defined data flow to convert raw data into a unified format of data as input for the the following model component:  raw data $\mapsto$ \textit{Preprocessor} $\mapsto$ \textit{Dataset} $\mapsto$ \textit{Dataloader} $\mapsto$ processed data. 

We display the statistics of all built-in datasets in Table~\ref{tab:dataset}.
As we can see, raw datasets vary in formats and features, so we first preprocess these raw datasets and convert them to a unified format. 
In \textit{Preprocessor}, we first tokenize input text by a tokenizer, extract numbers from the tokenized text by some simple rules, and record extracted numbers and map them into position-aware special tokens (a.k.a, number mapping).
To avoid infinite generation space in target, we convert equations into equation templates by replacing numbers with position-aware special tokens from number mapping. 
We add another special token $<bridge>$ for the multiple equations generation task to convert the equation forest to a tree. 
Hence it can be treated as the single equation generation task. 
Note that different models require us to prepare different data formats and features. 
For example, Bert-based MWP models use WordPiece embeddings~\cite{wu2016googles} instead of word embeddings.
For another example, Graph2tree models utilize external information, like the results of the dependency parser, to construct the graph.
Hence we customize the preparation of data preprocessing after basic preprocessing. 
Users can add a new dataset to our framework by referring to our processing step.

We design the \textit{Dataset} module to do data preparation. 
The design of \textit{AbstractDataset} is to include some shared attributes and basic functions. 
Any specific dataset class or user customized dataset class can inherit \textit{AbstractDataset} with few modifications. 

After the \textit{Dataset} module, \textit{DataLoader} module selects features from the processed data to form tensor data (PyTorch), which can be directly used in the model component. 
\textit{AbstractDataLoader} class, including common attributes and basic functions, allows users to easily create new \textit{DataLoaders} for new models.

\subsection{Model Component}

\input{Tables/models}

We organize the implementations of MWP solving methods in the model component. 
The objective of the model component is to disentangle model implementation from data processing, evaluation, execution, and other parts, which benefits users to focus on the model itself. 
We unify the implementation of a model.
Specifically, we provide three interface functions for loss calculation, prediction, and test, respectively. 
When users deploy or add a new model with \texttt{MWPToolkit}, they can simply focus on these interface functions without considering other parts. 
Such a design enables users to develop new algorithms easily and quickly. 
Besides, the commonly-used components of the implemented models have been decoupled and shared across different models for code re-usage.

We have carefully surveyed the recent literatures and selected the commonly-used MWP solving models in our library. 
As the first released version, we have implemented 17 MWP solving models in the four categories: \textit{Seq2seq}, \textit{Seq2tree}, \textit{Graph2tree}, and \textit{Pretrained Language Models}. 
In the future, more methods will be added into our toolkit as the regular update, like MathDQN~\cite{wang2018mathdqn}, EPT~\cite{kim2020ept}, KAS2T~\cite{wu2020ka-s2t}, and NumS2T~\cite{wu2021nums2t}. 
We summarize all the implemented models in Table~\ref{table:model}. 

For all the implemented models in \texttt{MWPToolkit}, we ensure that these re-implementations are correct and that the experiments by our framework are reliable.
We set the same hyper-parameters as the ones in original papers and ensure the re-implemented result should be approximate to the reported result.
The detailed performance comparison between our re-implementation and original results is shown in Table~\ref{tab:single} and Table~\ref{tab:multiple}.

\subsection{Evaluation Component}

Our toolkit standardizes the evaluation of MWP solving models with \textit{Equation Accuracy} and \textit{Answer Accuracy} for single equation generation or multiple equations generations. 
Equations accuracy is computed by measuring the exact match of predicted equations and ground-truth equations. 
For answer accuracy, we first check the validation of predicted equations. 
The answer accuracy is $0$ if the predicted equations are invalid or unsolvable. 
We then calculate the answer using our encapsulated calculation module and compare it with the ground-truth answer. 
If their difference is less than $1e-5$, we regard it as $1$ and $0$ otherwise. 


\subsection{Execution}
On the top of above components, we implement training and testing paradigms, where two options are provided.
One is to follow the standard train-test splitting if the splitting data is given in the original dataset.
Another is to conduct $k$-fold cross-validation.
To improve the performance, we also implement a series of hyper-parameter search strategies, such as beam search, greedy search, sampling strategy.

\section{Usage of \texttt{MWPToolkit}}

This section shows how users run the existing models and incorporate new models with our toolkit.

\subsection{Running Existing Models}

\begin{figure*}[t] 
\centering
\includegraphics[width=1.0\linewidth]{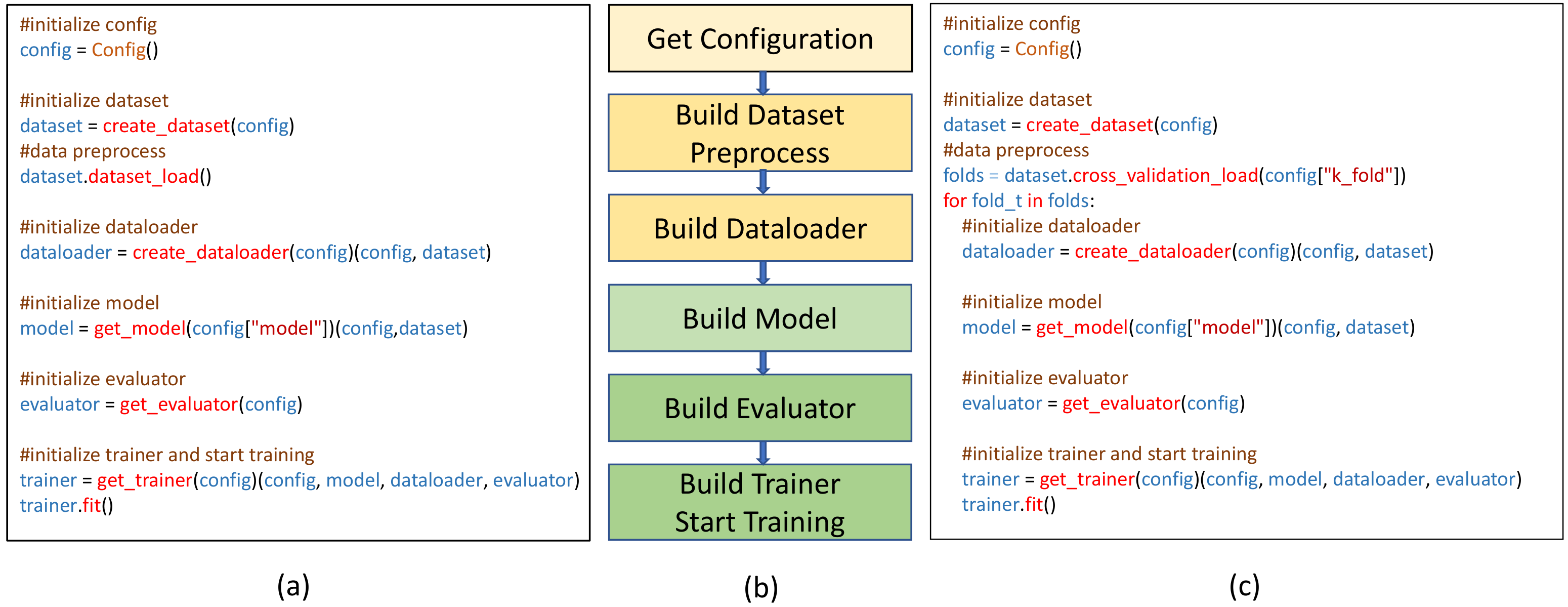} 
\vspace{-20px}
\caption{Examples of how to use our \texttt{MWPToolkit}.
Figure (a) illustrates the code of running models using the train-test split. Figure (b) is about the usage flow of the toolkit. Figure (c) shows the code of running models using $k$-fold cross-validation.
}
\label{fig:flow} 
\vspace{-10px}
\end{figure*}

Figure~\ref{fig:flow} shows the procedure of running an existing model in \textit{MWPToolkit}. 
Firstly, users need a configuration file to set up the experimental environment. In the configuration file, users should specify an existing model, a dataset, a task, and other hyper-parameters regarding the model and training. 
The class \textit{Configure()} loads all information on configuration for the subsequent steps. 
Then, the toolkit preprocesses data and organizes the dataset by calling the function \textit{create\_dataset()}. 
Based on the processed dataset, users can use the function \textit{create\_dataloader()} to convert data to the tensor format for training, validation, and test with the specified batch size and other hyper-parameters like the maximum length of the input sequence. 
Later, the function \textit{get\_model()} is used to get access to the model that the users would like to run. 
Next, users can employ the function \textit{get\_trainer()} to build an executable MWP solver based on the dataloader, model, and specified task obtained in previous steps. 
Eventually, users run the function \textit{trainer.fit()} to start the training and evaluation.

\subsection{Developing New MWP Sovlers}

\texttt{MWPToolkit} is an extensible and easy-to-use framework. 
It is convenient for users to add a new MWP solving model or a new benchmark dataset into \texttt{MWPToolkit} by filling up the specified interfaces. 
In the following, we present the details of how to add a new dataset and model.

\subsubsection{Add a New Dataset}

To add a new dataset, users need to inherit the abstract class \textit{AbstractDataset} and are required to fill in the functions: \textit{\_\_init\_\_()}, \textit{\_load\_data()}, \textit{\_preprocess()}, and \textit{\_build\_vocab()}.  
\textit{\_\_init\_\_()} is used to set up parameters of the dataset. 
\textit{\_load\_data()} is used to load the entire raw dataset or split training, validation, and test sets. 
The function \textit{\_preprocess()} is used to process the raw dataset and prepare the processed dataset for later usage in other modules. 
\textit{\_build\_vocab()} is used to build shared or separate vocabulary dictionaries for the encoder and decoder. 
The users can fill in the above required implemented functions to create a customized dataset class quickly.

\subsubsection{Add a New Model}

To add a new model, users need to complete three functions in a new model class: \textit{\_\_init\_\_()}, \textit{calculate\_loss()}, and \textit{model\_test()}. 
\textit{\_\_init\_\_()} is used to build the model and initialize the parameters. \textit{calculate\_loss()} is used to calculate the loss for training based on the model prediction and ground-truth equations. 
\textit{model\_test()} prepares suitable evaluation protocols and is executed for evaluation of model performance on a specified dataset and task.

\section{Performance Comparison}

To evaluate the models in \texttt{MWPToolkit}, we conduct extensive experiments to compare 17  MWP solving models on 4 widely-used single equation generation benchmark datasets and 2 multiple equations generation benchmarks.
In our experiments, if models have been evaluated on certain datasets, we run models with the parameter configurations described in their original papers. 
Otherwise, we run hyper-parameter search to search a group of hyper-parameters for these models. 
In the following sections, we discuss the detailed performance comparison.

\subsection{Single Equation Generation}

Table~\ref{tab:single} displays the results of models on single equation generation datasets.
We include four datasets for the single equation generation task, i.e., Math23k, MAWPS-s, ASDiv-a, and SVAMP. We can see from Table~\ref{tab:single}. 
We report three types of results for a model on a dataset. 
The first two columns are equation accuracy (Equ. Acc) and answer accuracy (Ans. Acc). The third column is the results reported in the original papers under their settings, such as train-test split (* means train-test split) or 5-fold cross-validation. 
Note that for any models, the results based on 5-fold cross-validation are less than those based on train-test split because the number of training examples of 5-fold cross-validation is smaller than those in train-test split. 
As shown in table~\ref{tab:single}, answer accuracy in the $k$-fold cross-validation setting by our \texttt{MWPToolkit} are either better than original answer accuracy or close to them. 
Through our experiments, we can observe that Graph2Tree and RoBERTaGen are the most effective baselines, which means it is potential for researchers to develop better models based on these two model categories.

\input{Tables/single_equation}

\subsection{Multiple Equations Generation}

\input{Tables/multiple_equations}

We add another special token $<bridge>$ to convert equation forest to a tree for the multiple equations generation task. 
Hence it can be treated as the single equation generation task. 
We apply the 17 models on two multiple equation generation datasets, i.e., DRAW1K and HMWP. Their results are shown in Table~\ref{tab:multiple}. 
As we can observe in table~\ref{tab:multiple}, to our surprise, LSTMVAE achieves the best performance on Draw1K, and GPT-2 achieves the best performance on HMWP. 
Most researchers focus on improving performance on the single equation generation task, while few researchers develop models on the multiple equation generation task in the MWP solving community. 
We hope the results shown in table~\ref{tab:multiple} can help researchers develop more powerful and effective models for solving MWPs with multiple equations as their generation targets.

\section{Related Work}

In NLP community, there have been a number of toolkits that managed to summarize the existing methods and establish a unified framework for a certain task, such as OpenNMT~\cite{klein2017opennmt} and TexBox~\cite{li2021textbox} for text generation tasks, ExplainaBoard~\cite{liu2021explainaboard} for evaluating interpretable models, Photon~\cite{zeng2020photon} for text-to-SQL tasks, and Huggingface's Transformers~\cite{wolf2020transformers} for model pretraining. 
To our best knowledge, there is no such a unified and comprehensive framework for MWPs solving task.
Therefore, we release \texttt{MWPToolkit}, which includes a considerable number of benchmark  datasets and deep learning-based solvers. 

Recently, a large number of new MWP solving methods have been proposed, including graph neural networks based ~\cite{li2020graph2tree}, template based~\cite{lee2021template}, neural symbolic~\cite{qin2021symbolic}, pre-trained based~\cite{liang2021mwp}, multilingual pre-trained based methods~\cite{tan2021multilingual}, and solvers using external information and signals~\cite{liang2021teacher, wu2021nums2t}. In addition, weakly supervised learning for MWP solving~\cite{hong2021lbf, chatterjee2021weakly} and supervised learning for geometric problem solving~\cite{lu2021gps, chen2021geoqa} have recently attracted much researchers' attention. More work on math word and geometric problem solving can be found in the survey paper~\cite{zhang2019mwpsurvey}. We will update more above methods to the toolkit in the future.

\section{Conclusion}

This paper presented an extensible, modularized, and easy-to-use toolkit, \texttt{MWPToolkit}, the first open-source framework for solving MWPs. 
In our \texttt{MWPToolkit}, we decompose the procedure of existing MWP methods into multiple components and decouple their models into highly reusable modules. 
We also provide a hyper-parameter search function for a fairer comparison. 
Furthermore, we implement and compare 17 MWP solving models, including Seq2Seq, Seq2tree, Graph2Tree, and commonly-used non-pretrained models and pretrained models, on 4 widely-used single equation generation benchmark datasets and 2 multiple equations generation benchmarks. These features enable our \texttt{MWPToolkit} to be suitable for researchers to reproduce reliable baseline models and develop new MWP solving methods quickly.

In the future, we will continue to add more benchmark datasets, the latest published MWP solvers, and commonly-used models into \texttt{MWPToolkit} as the regular update. We welcome more researchers and engineers to join, develop, maintain, and improve this toolkit to push forward the development of the research on MWPs solving.

\section*{Acknowledgement}

The authors would like to thank everyone who has contributed to make MWPToolkit a reality. Thanks to the TextBox~\cite{li2021textbox}, CRSLab~\cite{zhou2021crslab}, and RecBole~\cite{zhao2020recbole} for such elegant and easy-to-use libraries. We refer to these libraries and learn a lot from them.

\bibliographystyle{acl_natbib}
\bibliography{acl2020}

\end{document}

%% file: Tables/example.tex
\begin{table}[!ht]
\small
\begin{tabular}{p{7cm}}
\toprule 

\textbf{Single Equation Generation}: 

Paco had 26 salty cookies and 17 sweet cookies. He ate 14 sweet cookies and 9 salty cookies. How many salty cookies did Paco have left?\\
 \textbf{Equation}: 
 
 $x = 26 - 9$

 \textbf{Answer}: 
 
 $x = 17$\\
 
 \midrule
 
 \textbf{Multiple Equations Generation}: 
 
 Jerome bought 12 CDs. Some cost 7.50\$ each, and the rest cost 6.50 each . How many CDs were bought at each price if he spent 82 dollars? \\
 
 \textbf{Equation}:
 
 $7.5 \times x+6.5 \times y=82, $ \\
 $x+y=12$

 \textbf{Answer}: 
 
 $x=4, y=8$\\
 
 \bottomrule
\end{tabular}
\caption{Two examples of Math Word Problems. 
We display single equation generation and multiple equations generation.
}
\label{tab:example}
\vspace{-10px}
\end{table}

%% file: Tables/datasets.tex
\begin{table*}[t]
\centering
\resizebox{1.0\textwidth}{!}{
\begin{tabular}{lcccccc}
\toprule
Dataset & Language &  Task  & \# Examples & \# Multi-Equ & Hard Set &  Reference \\
\midrule
MAWPS-s~ & en & Single equation generation  & 1,987 & - & - & \cite{Koncel2016mawps}\\
Draw1K~ & en & Multiple equations generation  & 1,000 & 745 & - & \cite{Upadhyay2017AnnotatingDA} \\
Math23K~ & zh & Single equation generation  & 23,162  & - & - & \cite{wang2017dns}\\
HMWP~ & zh & Multiple equations generation & 5,491& 1,789& - & \cite{qin2020sau-solver} \\
ASDiv-a~ & en & Single equation generation & 1,238 & - & - & \cite{Miao2020ADC} \\
SVAMP~ & en & Single equation generation  & 3,138 &- & 1,000 & \cite{Patel2021AreNM} \\
\bottomrule
\end{tabular}
}
\caption{
The collected datasets in \texttt{MWPToolkit}. ``\# Multi-Equ'' stands for the number of examples, the targets of which are multiple equations. 
``Hard Set'' means an external challenging or adversarial test set.
}
\vspace{-10px}
\label{tab:dataset}
\end{table*}

%% file: Tables/models.tex
\begin{table*}[t]
\centering
\resizebox{0.86\textwidth}{!}{

\begin{tabular}{l|c|c|c|c|c}
\toprule
Type &Model &Encoder &Decoder & Pretrained Model & Reference\\ \midrule
\multirow{7} * {Seq2Seq} &DNS & GRU & LSTM & - &\cite{wang2017dns} \\
 &MathEN & BiLSTM & LSTM & - &\cite{wang2018mathdqn}\\ 
 &Saligned & BiLSTM & LSTM & -&\cite{chiang2019saligned}\\ 
 &GroupATT & BiLSTM & LSTM & - &\cite{li2019groupatt}\\ 
\cline{2-6}
 &AttSeq2Seq & LSTM & LSTM& - &\cite{bahdanau2014neural}\\
 &LSTMVAE & LSTM & LSTM& - &\cite{zhang2016lstmvae}\\
 &Transformer & Transformer& Transformer  & - &\cite{vaswani2017attention}\\
 \midrule
 \multirow{5} * {Seq2Tree} &TRNN & BiLSTM & LSTM& - &\cite{wang2019trnn} \\
 &AST-Dec & BiLSTM & TreeLSTM & - &\cite{liu2019astdec}\\ 
 &GTS & GRU & TreeDecoder & -&\cite{xie2019gts}\\ 
 &SAU-Solver & GRU & TreeDecoder & - &\cite{qin2020sau-solver}\\ 
 &TSN & GRU & TreeDecoder & - &\cite{zhang2020teacher}\\ 
 \midrule
 \multirow{2} * {Graph2Tree} & Graph2Tree & LSTM+GCN & TreeDecoder & -&\cite{zhang2020graph2tree} \\
 &MulltiE\&D & GRU+GCN & GRU& - &\cite{shen2020multiED}\\ 
 \midrule
 
 \multirow{3} * {Pretrained based}  & BERTGen & BERT & Transformer & BERT &\cite{devlin2018bert} \\
 
 &RoBERTaGen  & RoBERTa & Transformer & RoBERTa &\cite{liu2019roberta} \\
 
 & GPT-2 & - & Transformer & GPT-2 &\cite{radford2019gpt2} \\

\bottomrule
\end{tabular}
}
\caption{
The implemented models in \textit{MWPToolkit}. Currently, the toolkit includes four types of models: Seq2Seq, Seq2Tree, Graph2Tree, and Pretrained models.
}
\label{table:model}
\vspace{-10px}
\end{table*}

%% file: Tables/single_equation.tex
\begin{table*}[!t]
\resizebox{1.0\textwidth}{!}{
\centering
\begin{tabular}{c|ccc|ccc|ccc|ccc} 
\toprule
\multirow{3}{*}{\textbf{Model}} & \multicolumn{12}{c}{\textbf{Datasets}}       \\ 
\cline{2-13}
      & \multicolumn{3}{c|}{\textbf{Math23K}} & \multicolumn{3}{c|}{\textbf{MAWPS-s}} & \multicolumn{3}{c|}{\textbf{ASDiv-a}} & \multicolumn{3}{c}{\textbf{SVAMP}}   \\ 
      & \textbf{Equ. Acc} & \textbf{Ans. Acc} & \textbf{OA Acc}  & \textbf{Equ. Acc} & \textbf{Ans. Acc} & \textbf{OA Acc} & \textbf{Equ. Acc} & \textbf{Ans. Acc} & \textbf{OA Acc} & \textbf{Equ. Acc} & \textbf{Ans. Acc} & \textbf{OA Acc}  \\ 
\hline
DNS&57.1 & 67.5 & 58.1 &78.9 &86.3 &59.5 &63.0 &66.2 &- &22.1 &24.2 &- \\ 

MathEN&66.7 &69.5 &66.7* &85.9 &86.4 &69.2 &64.3 &64.7 &- &21.8 &25.0 &- \\

Saligned& 59.1&69.0 &65.8 &\textbf{86.0} &86.3 &- &66.0 &67.9 &- &23.9 &26.1 &- \\ 

GroupAtt&56.7 &66.6 &66.9 &84.7 &85.3 &76.1 &59.5 &61.0 &- &19.2 &21.5 &- \\

\cline{2-13}

AttSeq& 57.1 & 68.7& 59.6 &79.4 &87.0 &79.7 &64.2 &68.3 &55.5 &23.0 &25.4 &24.2 \\

LSTMVAE& 59.0 &70.0 & - &79.8 &88.2 &- &64.0 &68.7 &- &23.2 &25.9 &- \\

Transformer&52.3 &61.5 &62.3* &77.9 &85.6 &- &57.2 &59.3 &- &18.4 &20.7 &- \\

\midrule

TRNN&65.0 &68.1 &66.9* &\textbf{86.0} &86.5 &66.8 &68.9 &69.3 &- &22.6 &26.1 &- \\

AST-Dec&57.5 &67.7 &69.0* &84.1 &84.8 &- &54.5 &56.0 &- &21.9 &24.7 &- \\

GTS&63.4 &74.2 &74.3 &83.5 &84.1 &82.6 &67.7 &69.9 &\textbf{71.4} &25.6 &29.1 &30.8 \\

SAU-Solver&64.6 &75.1 &74.8 &83.4 &84.0 &- &68.5 &71.2 &- &27.1 &29.7 &- \\

TSN&63.8 &74.4 &75.1 &84.0 &84.7 &\textbf{84.4} &68.5 &71.0 &- &25.7 &29.0 &- \\
\midrule

Graph2Tree& 64.9&75.3 &75.5 &84.9 &85.6 &83.7 &\textbf{72.4} &\textbf{75.3} &- &~\textbf{31.6} &~\textbf{35.0} & ~\textbf{36.5} \\

MultiE\&D&\textbf{65.5} &76.5 & \textbf{76.9} &83.2 &84.1 &- &70.5 &72.6 &- &29.3 &32.4 &- \\
\midrule

BERTGen&64.8 &76.6 &- &79.0 &86.9 &- &68.7 &71.5 &- &22.2 &24.8 &- \\

RoBERTaGen&65.2 &\textbf{76.9} &- &80.8 &\textbf{88.4} &- &68.7 &72.1 &- &27.9 &30.3 &- \\

GPT-2&63.8 &74.3 &- &75.4 &75.9 &- &59.9 &61.4 &- &22.5 &25.7 &- \\

\bottomrule

\end{tabular}
}
\caption{Performance comparisons of different methods on single equation generation task. ``Equ. Acc'' is equation accuracy. ``Ans. Acc'' stands for answer accuracy. ``OA Acc'' means original answer accuracy in previous papers. ``*'' means train-test split.}\label{tab:single}
\vspace{-15px}
\end{table*}

%% file: Tables/multiple_equations.tex
\begin{table}[!t]
\centering
\resizebox{0.5\textwidth}{!}{

\begin{tabular}{c|ccc|ccc} 
\toprule
\multirow{3}{*}{\textbf{Model}} & \multicolumn{6}{c}{\textbf{Datasets}}       \\ 
\cline{2-7}
      &  \multicolumn{3}{c|}{\textbf{Draw1K}} & \multicolumn{3}{c}{\textbf{HMWP}}   \\ 
       & \textbf{Equ. Acc} & \textbf{Ans. Acc} & \textbf{OA Acc} & \textbf{Equ. Acc} & \textbf{Ans. Acc} & \textbf{OA Acc}   \\ 
\hline
DNS&35.8 &36.8 &- &24.0 &32.7 &- \\ 

MathEN&38.2 &39.5 &- &32.4 &43.7 &- \\

Saligned&36.7 &37.8 &- &31.0 &41.8 &- \\ 

GroupAtt&30.4 &31.4 &- &25.2 &33.2 &- \\

\cline{2-7}

AttSeq&39.7 &41.2 &- &32.9 &44.7 &- \\

LSTMVAE&\textbf{40.9} &\textbf{42.3} &- &33.6 &45.9 &- \\

Transformer&27.1 &28.3 &- &24.4 &32.4 &- \\

\midrule

TRNN&27.4 &28.9 &- &27.2 &36.8 &- \\

AST-Dec&26.0 &26.7 &- &24.9 &32.0 &- \\

GTS&38.6 &39.9 &- &33.7 &44.6 &- \\

SAU-Solver&38.4 &39.2 &- &33.1 &43.7 &\textbf{44.8} \\

TSN&39.3 &40.4 &- &34.3 &44.9 &- \\
\midrule

Graph2Tree&39.8 &41.0 &- &34.4 &45.1 &- \\

MultiE\&D&38.1 &39.2 &- &34.6 &45.3 &- \\
\midrule

BERTGen&33.9 &35.0 &- &29.2 &39.5 &- \\

RoBERTaGen&34.2 &34.9 &- &30.6 &41.0 &- \\

GPT-2&30.7 &31.5 &- &\textbf{36.3} &\textbf{49.0} &- \\

\bottomrule

\end{tabular}
}
\caption{Performance comparison of different methods on multiple equations generation task. 
``Equ. Acc'' is equation accuracy. ``Ans. Acc'' stands for answer accuracy. ``OA Acc'' means original answer accuracy in previous papers.
}\label{tab:multiple}
\vspace{-15px}
\end{table}